# Combining Domain-Specific Models and LLMs for Automated Disease Phenotyping from Survey Data.

Gal Beeri[1], Benoit Chamot[1], Elena Latchem[2], Shruthi Venkatesh[2], Sarah Whalan[3], Van Zyl Kruger[1,] David Martino [2,3,4] *.

*1 DataDivers.io, 111 Colin Street, West Perth WA 6000.*
*2 The University of Western Australia, 35 Stirling Highway, Crawley, Perth, WA 6009.*
*3 The ORIGINS Project, The Kids Research Institute Australia, Nedlands, WA 6009, Australia*
*4 Wal-yan Respiratory Research Centre, The Kids Research Institute Australia, Nedlands, Perth, WA 6009, Australia*

**Running title:** Automated Phenotyping of Survey Data with GenAI

**Author contributions:** Conception and design of study: DM, VK. Data analysis and interpretation: GB, BC, SW. Manuscript drafting: EL, SV. Critical revisions: GB, BC, EL, SV, SW, VK, DM

**Funding and support:** The Generative AI Challenge is funded by grants from the Future Health Research and Innovation Fund (FHRIF), Grant ID IC2023-GAIA/11.

**Conflict of interest statement:** The authors declare no conflicts of interest.

* **Correspondence:** David Martino, david.martino@thekids.org.au, The Kids Research Institute Australia, 15 Hospital Avenue Nedlands, 6009, Western Australia.

**Abstract**

This exploratory pilot study investigated the potential of combining a domain-specific model, BERN2, with large language models (LLMs) to enhance automated disease phenotyping from research survey data. Motivated by the need for efficient and accurate methods to harmonize the growing volume of survey data with standardized disease ontologies, we employed BERN2, a biomedical named entity recognition and normalization model, to extract disease information from the ORIGINS birth cohort survey data. After rigorously evaluating BERN2's performance against a manually curated ground truth dataset, we integrated various LLMs using prompt engineering, Retrieval-Augmented Generation (RAG), and Instructional Fine-Tuning (IFT) to refine the model's outputs. BERN2 demonstrated high performance in extracting and normalizing disease mentions, and the integration of LLMs, particularly with Few Shot Inference and RAG orchestration, further improved accuracy. This approach, especially when incorporating structured examples, logical reasoning prompts, and detailed context, offers a promising avenue for developing tools to enable efficient cohort profiling and data harmonization across large, heterogeneous research datasets.

**Introduction**

The increasing availability of research survey data from cohort studies and clinical trials offers unprecedented opportunities to advance biomedical research and improve healthcare (1). However, the lack of standardized ontologies for describing health characteristics and outcomes hinders data interoperability and harmonization across studies. Manual classification of survey data into consistent phenotype and disease ontologies is time-consuming, labour-intensive, error-prone, and costly, posing a significant bottleneck, considering the large amount of data available.

Natural Language Processing (NLP), a branch of artificial intelligence, offers promising solutions to automate this process. NLP techniques, such as Named Entity Recognition (NER) and Named Entity Normalisation (NEN), can extract and standardize disease mentions from diverse survey data, including structured and unstructured text formats (2). Recent advancements in NLP models, particularly those tailored for biomedical text, have shown potential for automating the curation of research survey data (3). However, existing approaches have limitations. Studies utilizing general-purpose LLMs like ChatGPT for biomedical NLP tasks have shown limited success (4), often being outperformed by purposed-built baseline models (5). While domain-specific pretraining has demonstrated advantages (6), the integration of specialized models like BERN2 (7), designed for NER and NEN tasks in biomedical text, remains underexplored.

This was an exploratory pilot study investigating the potential of combining a domain-specific model, BERN2, with large language models (LLMs) to enhance the efficiency of disease phenotyping from research survey data. We focused on applications that could automatically tag research data sets collected as part of the ORIGINS birth cohort of 10,000 West Australian families focused on understanding the early life determinants of non-communicable disease (8). Our use-case was to tag de-identified ORIGINS datasets administered as REDCap surveys documenting lifestyle, environment, and child health data with standardized disease ontologies from the MeSH database. MeSH is a comprehensive controlled vocabulary thesaurus developed and maintained by the National Library of Medicine. It provides a hierarchical structure of medical and health-related terms, allowing for consistent and precise tagging of research data. Our objective was to develop approaches that could automatically and accurately identify and categorize disease information within the diverse research datasets collected by the project. Developing such approaches could enable identifying and monitoring disease trends, facilitating targeted research in risk factors and disease relationships, enabling cross-cohort harmonisations, and enhancing findability and interoperability of cohort research data.

Our novel approach leveraged BERN2's capabilities in NER and NEN, utilizing BioBERT (9) for disease mention recognition and employing neural network-based normalisation methods. We adopted an iterative process of automated phenotyping using BERN2, followed by manual annotation of a subset of records to create a ground truth dataset. This dataset was used to rigorously evaluate BERN2's performance using various metrics, including accuracy, precision, recall, and F1 score. Furthermore, several LLMs were integrated to assess and potentially refine BERN2's outputs. Various prompt engineering techniques, such as few-shot inference (FSI) and Chain-of-Thought (CoT) prompting, were employed to optimize LLM performance. Additionally, augmented knowledge methods, including Retrieval-Augmented Generation (RAG) and Instructional Fine-Tuning (IFT), were explored to enhance the LLMs' contextual understanding and normalisation capabilities.

This paper provides a detailed analysis of these approaches, examining the strengths and limitations of different AI models, prompting techniques, and augmented knowledge methods. The findings demonstrate the potential of combining domain-specific models with LLMs to improve the accuracy and efficiency of automated disease phenotyping from research survey data.

**Methods**

**Data acquisition and processing**

The ORIGINS birth cohort comprises longitudinal survey data collected via REDCap from over 10,000 families, encompassing information across multiple domains, including demographics, lifestyle, environment, and health (8). For this study, a subset of 57,785 survey records representing unique questions posed to participants was extracted for analysis. The number of unique question-answer pairs within this subset was substantially larger due to the inclusion of free-text responses, which inherently exhibit significant variability. To facilitate automated phenotyping, the extracted data was de-identified and centralized into a data lake hosted in Amazon Web Services (AWS) S3. Pre-processing techniques were employed to transform the data into a machine-readable format, including stemming, padding, expanding acronyms, converting text to lowercase, and correcting spelling and punctuation errors. This study protocol was approved by the Ramsay Health Care (RHC) WA | SA Human Research Ethics Committee (HREC reference number 2023/ETH/0088).

**Automated Phenotyping Pipeline**

BERN2 was selected for automated phenotyping due to its demonstrated high accuracy in biomedical named entity recognition and normalisation. It is specifically optimized for speed and focuses on disease entities, aligning with the study goals (7). However, the original BERN2 codebase, designed for sequential processing of single text inputs, required modification to handle the large volume of survey data in this project.

Adaptations to the BERN2 model included enabling batch processing of data extracted from REDCap and running the model in an efficient, programmatic way. This involved stripping the code down to only the necessary functions for the ORIGINS project to improve efficiency, allowing the model to run independently from the web interface that was developed for it, and changing the input data to a format that is consistent with the RedCap data structure. The code was also adapted to run in the AWS SageMaker cloud environment used for the project. Furthermore, the output format was restructured to facilitate subsequent data analysis. These adaptations enabled efficient and accurate processing of the unique characteristics of the ORIGINS dataset.

BioBERT was employed as the NER model within BERN2 to identify disease mentions. Three distinct normalisation approaches were evaluated in forerunner experiments: BioSyn, sieve, and hybrid. The BioSyn neural network was implemented for this project as it outperformed sieve and hybrid in qualitative analysis. We leveraged only the disease entity NER functions to align with the project goals and normalised to the Medical Subject Headings (MeSH) medical ontologies which, while not as detailed as other ontology databases, utilises a highly structured vocabulary, reducing ambiguity.

**Manual Annotation of Survey Records**

To assess the accuracy of the automated phenotyping pipeline, a subset of 100 survey records was manually annotated. The outputs generated by the modified BERN2 model were imported into Doccano, an open-source annotation software (10). Two independent researchers then manually reviewed and tagged these records, assigning the corresponding concepts from the MeSH ontology, used in the BERN2 model, to each identified disease mention. Any discrepancies in annotations between the two researchers were resolved through adjudication by a third senior researcher. The 100 records selected for manual annotation were carefully curated to ensure a representative sample of the dataset. The sample included an equal proportion of records with free-text questions where disease mentions were expected and unexpected. This ensured that the evaluation encompassed a variety of response types and question formats, including slider, descriptive, binary, ratio dropdown, and checkbox questions. This diverse sample enabled a comprehensive assessment of the model's performance across different question categories and response types.

**LLM Integration**

This study explored the potential of Large Language Models (LLMs) to enhance the accuracy and efficiency of automated phenotyping. Several general-purpose LLMs were utilized, including Mistral 7B, Mixtral 8x 7B, Llama 2 7B, Llama 3 8B, and Llama 3 70B.

LLMs were employed in two primary ways:

1. BERN2 Output Evaluation: Given their superior context-awareness compared to BERN2, LLMs were used to assess the accuracy of BERN2's disease normalisation by analysing the preceding questions within each survey record. This helped determine whether BERN2 had assigned the correct ontology term, considering the broader context of the respondent's answers.
2. Chained LLMs for Validation: LLMs were 'chained' together in a sequential process where one LLM generated an output (e.g., a proposed disease concept) and another LLM evaluated the validity and appropriateness of that output. This approach aimed to leverage the strengths of different LLMs for improved accuracy.

<u>Fine-tuning Attempts</u>

To further enhance the LLMs' performance on this task, we attempted to fine-tune them using the ORIGINS survey data (11). This was done because the ORIGINS data provided the most specific and relevant context for our research question. While other data sources, such as

NCBI (12) were considered, they were deemed insufficiently specific for fine-tuning the LLMs to effectively extract context and answer user queries related to the ORIGINS survey content.

LLM Implementation Activities

Various techniques were explored to optimize the interaction with and output from the LLMs:

- Few-Shot Inference (FSI)(13): FSI was applied with 5-shot inference to enhance the accuracy of LLM outputs. One-shot, two-shot, and three-shot prompts were applied serially, and the outputs were evaluated using evaluation metrics (detailed below).
- Chain-of-Thought (CoT) Prompting: CoT prompting was employed to enhance the LLMs' reasoning capabilities by encouraging them to generate intermediate reasoning steps (14). Prompts incorporating phrases like "Let's think step by step" were used. Four different CoT prompts were tested on Llama 3 8B and Llama 3 70B, with each prompt paired with user inputs and one example. The LLMs were tasked with determining whether they agreed or disagreed with the concept normalised by BERN2. The LLMs' agreement, along with BERN2's outputs and the ground truth dataset, were compared to assess the LLMs' ability to improve BERN2's performance.
- Retrieval-Augmented Generation (RAG) and CoT: This approach combined zero-shot CoT prompting with RAG orchestration (15) allowing LLMs to access and utilize information from the MeSH ontology during the generation process. This aimed to improve reasoning, reduce hallucinations, and enhance the accuracy of LLM outputs. Four prompts (reference (FSI), simple CoT, strong CoT, and hybrid (FSI and CoT)) were tested on Llama 3 8B and Llama 3 70B.
- Instructional Fine-Tuning (IFT): Mixtral 8x 7B and Llama 3 8B were fine-tuned using IFT to adapt them to the specific task of disease normalisation (16). This involved augmenting the models with contextual knowledge relevant to the ORIGINS dataset.
- Retrieval Augmented/Aware Fine-Tuning (RAFT): RAFT was explored as a method to incorporate domain knowledge into the LLMs while improving RAG performance (17). This process involved preparing training data where each data point included: N questions (Q*), N distractor documents (Di) containing false information to challenge the retrieval of relevant content, an oracle document (D*) with accurate information, and corresponding chain-of-thought (CoT) answers (A*) to provide reasoning for the correct responses.
- FSI and RAG: This combined approach aimed to leverage the strengths of both FSI (providing structural framework) and RAG (enriching content with external knowledge) to reduce hallucinations and improve the specificity and relevance of LLM outputs. The process involved querying the LLM with FSI to determine whether a mention matched the disease concept normalised by BERN2, using MeSH documents stored in the vector database. Each document contained the concept name, concept ID, disease description, and a list of synonyms. Further, FSI was again leveraged to generate a summary to be compared with a predefined gold standard summary to determine the coherence score between them.
- **RAG FSI with Binary Flags:** Binary flags were used to control the model's use of external retrieval (RAG) and few-shot examples (FSI). This provided flexibility in how LLMs approached different tasks, allowing for a more tailored response depending on the query and available resources.

**Evaluation Metrics**

The performance of the AI models, primarily BERN2 and the integrated LLMs, was rigorously evaluated using a combination of metrics. The primary benchmark was the manually annotated ground truth (GT) dataset, enabling a direct comparison between the model outputs and human expert annotations.

To ensure accurate assessment, specific criteria were established for determining agreement between the model outputs and the GT dataset:
- **Mention Agreement:** Agreement for a disease mention required identical start and end indices within a given record. Even partial overlaps (e.g., "asthma" vs. "asthma episodes") were considered disagreements to maintain a strict evaluation standard.
- **Concept Agreement:** Agreement for a disease concept required an exact match of the concept ID between the model output and the GT dataset for a given mention.

Standard confusion matrix metrics were employed to quantify model performance:
- **True Positive (TP):** A mention correctly identified and normalised by the model.
- **True Negative (TN):** A mention correctly not identified by the model (i.e., a non-disease mention).
- **False Positive (FP):** A mention incorrectly identified as a disease by the model.
- **False Negative (FN):** A disease mention missed by the model.

These metrics were used to calculate the following performance indicators:
- **True Positive Rate (Recall):** TP / (TP + FN)
- **False Positive Rate:** FP / (TN + FP)
- **True Negative Rate:** TN / (TN + FP)
- **False Negative Rate:** FN / (TP + FN)
- **Precision:** TP / (TP + FP)
- **F1 Score:** 2 × (Precision × Recall) / (Precision + Recall)
- **Accuracy:** (TP + TN) / (TP + TN + FP + FN)

Beyond confusion matrix metrics, the following evaluation measures were also employed:

- **ROUGE-n:** This metric assessed the overlap of n-grams between the model-generated summaries and the reference summaries, providing insights into the quality and conciseness of the generated text.
- **Average Cosine Similarity:** This metric measured the coherence of the generated text by calculating the cosine similarity between sentence vector embeddings. Higher scores (representing smaller angles or differences between the vectors) indicate greater coherence and readability.

**Results**

**BERN2 Performance**

Table 1 presents the performance of BERN2 against the manually annotated ground truth dataset. The evaluation considered both mention-level recognition and concept-level normalisation. As shown in Table 1, BERN2 achieved high accuracy in both disease mention recognition and concept normalisation. The model demonstrated an F1-score of 87.9% for mention recognition and 83.1% for concept normalisation, indicating its effectiveness in identifying and standardizing disease information within the survey responses. Furthermore, the model exhibited a high true positive rate and a low false positive rate, suggesting its ability to accurately identify disease mentions while minimizing incorrect classifications.

**LLM Evaluation**

To further enhance the accuracy of automated phenotyping, various LLMs were integrated with BERN2, and their performance was assessed using different prompting techniques and knowledge augmentation methods.

<u>Zero-Shot Prompting</u>

Table 2 presents the results of using LLMs with zero-shot prompting. Zero-shot prompting yielded mixed results, with performance varying significantly across the LLMs and prompt types.

The Mixtral 8x 7B model demonstrated the highest accuracy (83.3%) when using the Concept vs. Concept prompt, suggesting its potential for improving disease normalisation. Mistral 7B achieved moderate accuracy (71.2%) with the same prompt type. When using the Concept vs. Mention prompt, both models showed a decrease in accuracy, with Mixtral 8x 7B achieving 76.5% and Mistral 7B only reaching 58.5%. Notably, the Concept vs. Concept prompt elicited a higher hallucination rate (28.8% for Mistral 7B and 16.7% for Mixtral 8x 7B) compared to the Concept vs. Mention prompt, where hallucination rates were not recorded.

This analysis indicates that the choice of LLM and prompt type significantly influences the effectiveness of zero-shot prompting for disease normalisation. While some LLMs show promise, careful selection and evaluation are crucial for optimizing performance and minimizing hallucination.

**Fine-tuned Models**

Fine-tuning the LLMs on the ORIGINS survey data yielded minimal improvements in performance. As shown in Table 3, the fine-tuned Mixtral 8x 7B model with 4 classes achieved an F1-score of 80% for BERN2 alignment and 88.9% for ground truth (GT) alignment, compared to 19.3% and 20.97%, respectively, for the baseline Mixtral 8x 7B model. However, increasing the number of classes to 10 resulted in a significant decrease in performance for both Mixtral 8x 7B and Llama 3 8B models. The fine-tuned Llama 3 8B model exhibited extremely low F1-scores (6.78% for BERN2 alignment and 6.25% for GT alignment). Incorporating Retrieval Augmented Generation (RAG) with fine-tuning led to marginal improvements for the Mixtral 8x 7B model but did not significantly enhance the performance of the Llama 3 8B model. These results suggest that fine-tuning on the ORIGINS survey data alone may not be sufficient for effectively adapting the LLMs to this specific task. The limited improvement observed with fine-tuning, particularly when using a higher number of classes, indicates the need for further exploration of alternative fine-tuning strategies, different hyperparameter settings, a more extensive training set, or the use of larger, more powerful LLMs.

**Few-Shot Inference with Retrieval-Augmented Generation**

Table 4 presents the results of combining few-shot inference (FSI) with Retrieval-Augmented Generation (RAG). Combining FSI with RAG led to substantial improvements in disease normalisation accuracy and coherence. The Llama 3 8B model achieved the highest F1-score (97%), precision (97%), and recall (97%) for ROUGE-1, indicating a high degree of overlap between the model-generated summaries and the reference summaries. Llama 3 8B also exhibited the highest coherence score (90%), suggesting that the generated text was well-structured and easy to understand. Both Llama 3 70B and Mixtral 8x 7B also demonstrated improvements with RAG FSI, albeit to a lesser extent than Llama 3 8B. These results highlight the potential benefits of providing LLMs with both structured examples (FSI) and access to external knowledge sources (RAG) for enhancing disease normalisation accuracy and generating coherent summaries.

**Chain-of-Thought Prompting**

Contrary to expectations, Chain-of-Thought (CoT) prompting did not improve the performance of the LLMs in this task. As shown in Table 5, the Llama 3 8B model achieved the highest normalised performance (1.00) and true positive rate (82.60%) when no CoT prompting was used. Incorporating simple, strong, or hybrid CoT prompts led to a decrease in both normalised performance and true positive rate, while increasing the false-negative rate. Similarly, for the Llama 3 70B model, the highest normalised performance (0.91) and true positive rate (75.00%) were observed with the strong CoT prompt, but these values were still lower than those achieved by the Llama 3 8B model without CoT. These results suggest that CoT prompting may not be effective for enhancing the performance of LLMs in this specific task. The decrease in performance observed with CoT prompts could be attributed to several factors, including the relatively small size of the LLMs used (below the 100B-parameter threshold recommended for CoT [cite Wei et al.]) and the potential for CoT to introduce confusion in the reasoning process, especially for smaller models. While CoT was not further explored in this study, it could be revisited with larger or fine-tuned LLMs in future research.

**Embeddings**

The use of embeddings, which incorporate semantic information into the LLM's vectorized representation of the text, had a mixed impact on disease normalisation accuracy. As shown in Table 6, the default embedding achieved the highest ROUGE-1 F1-score (66%), precision (72%), and recall (64%), indicating a good level of overlap between the model-generated summaries and the reference summaries. The default embedding also resulted in the highest coherence score (84%), suggesting that the generated text was better-structured or easier to understand compared to other embedding methods. The PubMed embedding led to the lowest ROUGE-1 scores across all metrics, while the Jina embedding showed moderate performance on ROUGE-1 but the lowest coherence score (75%). These results suggest that the choice of embedding method can influence both the accuracy and coherence of LLM-generated summaries for disease normalisation, with default embeddings outperforming alternative embeddings. While incorporating semantic information through embeddings can be beneficial, careful selection and evaluation of embedding methods are crucial for optimizing performance.

**Discussion**

This study investigated the potential of combining a domain-specific model, BERN2, with LLMs to enhance automated disease phenotyping from research survey data. Our findings demonstrate that BERN2 achieves high accuracy in both disease mention recognition and concept normalisation, supporting its suitability for extracting and standardizing disease information from complex survey responses.

The integration of LLMs yielded mixed results. Zero-shot prompting showed promise with certain LLMs and prompt types, but careful selection and evaluation are crucial to optimize performance and minimize hallucination. Fine-tuning the LLMs on the ORIGINS survey data resulted in only marginal improvements, suggesting limitations in the data's size or diversity for effectively adapting LLMs to this specific task. Combining few-shot inference (FSI) with Retrieval-Augmented Generation (RAG) led to substantial improvements in disease normalisation accuracy and coherence, highlighting the benefits of providing LLMs with both structured examples and access to external knowledge sources. However, Chain-of-Thought (CoT) prompting did not enhance performance, possibly due to the relatively small size of the LLMs used or the potential for CoT to introduce confusion in the reasoning process. The use of alternative embeddings reduced accuracy and coherence, emphasizing limited value in exploring alternative embeddings.

Our findings are consistent with previous research indicating the limitations of general-purpose LLMs like ChatGPT for biomedical NLP tasks (4), and the need for larger well annotated data sets (18). While domain-specific pretraining has shown advantages (9, 19, 20), this study demonstrates the potential of integrating specialized models like BERN2, designed for NER and NEN tasks in biomedical text. The combination of domain-specific models with LLMs, particularly when employing techniques like FSI with RAG, represents a novel approach for enhancing automated phenotyping.

This study has some limitations. The relatively small size of the manually annotated dataset may have limited the evaluation of certain techniques. Additionally, the computational resources required for fine-tuning and evaluating large LLMs posed challenges, leading to a limit of the number of fine-tuning steps that could be performed. Future research could explore the use of larger annotated datasets and more powerful LLMs to further investigate the potential of this approach. Despite these limitations, our adaptation of the BERN2 model to process large volumes of heterogenous research data, and exploration of various LLMs and prompt engineering techniques represent a novel and innovative approach to computational phenotyping, providing valuable insights into promising future approaches.

## Conclusion

This study demonstrates the potential of combining domain-specific models like BERN2 with LLMs, combining FSI prompt engineering with RAG to improve the accuracy and efficiency of automated disease phenotyping from research survey data. This approach offers a promising avenue for accelerating biomedical research and improving healthcare through efficient data harmonization and analysis. Future work will focus on refining LLM integration strategies and expanding the evaluation to larger datasets and more diverse phenotyping tasks.

**Acknowledgements:** We wish to acknowledge the support of Amazon AWS who were part of this project. Ellie Orton and James Bashforth provided valuable contributions into cloud architecture. We also wish to acknowledge the contributions of Michael James Brown, Pas Casalino, Sheldon Scott, Maddy Sommer and Conall Butler from the DataDivers.io team who provided in-kind support to the project. We wish to acknowledge the families involved in the ORIGINS cohort who contributed vital data sets for this study. We extend our sincere thanks to the following teams and individuals who have made The ORIGINS Project possible: The ORIGINS Project team; Joondalup Health Campus (JHC); members of ORIGINS Project Community Reference and Participant Reference Groups; Research Interest Groups and the ORIGINS Project Scientific Committee; Telethon Kids Institute; City of Wanneroo; City of Joondalup; and Fiona Stanley. Google Gemini was used to augment the clarity of the first written draft of this paper. Human oversight was maintained throughout, and all co-authors reviewed the final manuscript for accuracy.

Recognition. 2021 International Joint Conference on Neural Networks (IJCNN); 2021 18-22 July 2021.

# Tables

**Table 1 – BERN2 results against ground truth data set.**

| Task | NER (%) | | | | | | NEN (%) |
|---|---|---|---|---|---|---|---|
| | F1 | accuracy | true positive | true negative | false positive | false negative | accuracy |
| BERN2 | 87.9 | 83.1 | 93.2 | 64.3 | 35.7 | 6.82 | 79.2 |

*NER – Named entity recognition; NEN – Named entity normalisation.*

**Table 2 – Zero-Shot Prompting**

|  | correct answers (%) | hallucination rate (%) |
|---|---|---|
| Concept vs Concept Prompt |  |  |
|     Mistral 7B | 71.20 | 28.80 |
|     **Mixtral 8x 7B** | **83.30** | **16.70** |
| Concept vs Mention Prompt |  |  |
|     Mistral 7B | 58.50 | NR |
|     **Mixtral 8x 7B** | **76.50** | NR |

*NR = not recorded.*

**Table 3 – Model Performance After Fine-Tuning**

|  | BERN2 alignment (%) | | | | GT alignment (%) | | | |
|---|---|---|---|---|---|---|---|---|
|  | F1 | P | R | A | F1 | P | R | A |
| **FTM (Zero-Shot)** | | | | | | | | |
| Mixtral 8x 7B (BM) | 19.30 | 78.57 | 100.0 | 88.00 | 20.97 | 85.71 | 92.31 | 88.89 |
| Mixtral 8x 7B FTM (4 classes) | 80.00 | 75.00 | 85.71 | 63.64 | 88.89 | 100.0 | 80.00 | 83.33 |
| Mixtral 8x 7B FTM (10 classes) | 90.91 | 100.0 | 83.33 | 31.58 | 91.42 | 100.0 | 84.21 | 30.65 |
| Llama 3 8B FTM (10 classes) | 6.78 | 40.0 | 3.7 | 3.51 | 6.25 | 40.0 | 3.39 | 3.23 |
| Llama 3 8B BM (10 classes) | 3.45 | 25.0 | 1.85 | 1.75 | 3.17 | 25.0 | 1.69 | 1.61 |
| **FTM-RAG** | | | | | | | | |
| Mixtral 8x 7B (BM) | 29.51 | 50.00 | 20.93 | 17.31 | 47.22 | 65.38 | 36.96 | 30.91 |
| Mixtral 8x 7B FTM (10 classes) | 26.67 | 47.06 | 18.60 | 15.38 | 47.78 | 62.5 | 37.01 | 31.55 |

*FTM – Fine tune model; FTM-RAG – fine-tuned model, receiver augmented generation. GT = ground truth data set, P=precision, R=recall, A=accuracy.*

**Table 4 – Few-Shot Inference with Retrieval-Augmented Generation**

|  | ROUGE-1 | | | Coherence | BERN2 alignment accuracy | GT alignment accuracy |
|---|---|---|---|---|---|---|
|  | F1 | P | R |  |  |  |
| RAG FSI (2nd run) |  |  |  |  |  |  |
| Mixtral 8x 7B | 0.76 | 0.82 | 0.72 | 0.83 | 0.74 | 0.66 |
| Llama 3 70B | 0.88 | 0.88 | 0.89 | 0.75 | 0.86 | 0.77 |
| Llama 3 8B | 0.97 | 0.97 | 0.97 | 0.9 | 0.92 | 0.80 |

*GT = ground truth data set, P=precision, R=recall.*

**Table 5 – RAG Few-Shot Inference with Binary Flags**

|  | BERN2 alignment accuracy | GT alignment accuracy |
|---|---|---|
| RAG FSI with Binary Flags  Mistral 8x 7B | 0.70 | 0.63 |

**Table 6 – Results of Chain-of-Thought Prompting**

|  | normalised performance | true positive (%) | false negative (%) |
|---|---|---|---|
| Llama 3 8B |  |  |  |
| **No CoT** | **1.00** | 82.60 | 4.50 |
| Simple CoT | 0.98 | 81.10 | 6.10 |
| Strong CoT | 0.96 | 79.50 | 7.60 |
| Hybrid CoT | 0.87 | 72.00 | 15.20 |
| Llama 3 70B |  |  |  |
| No CoT | 0.71 | 58.30 | 28.80 |
| Simple CoT | 0.82 | 67.40 | 19.70 |
| Strong CoT | 0.91 | 75.00 | 12.10 |
| Hybrid CoT | 0.83 | 68.20 | 18.90 |

*CoT = chain of thought.*

**Table 7 – Embeddings**

| Embedding | ROUGE-1 | | | coherence |
|---|---|---|---|---|
| | F1 | P | R | |
| **Default** | **0.66** | **0.72** | **0.64** | 0.84 |
| PubMed | 0.57 | 0.65 | 0.54 | 0.76 |
| Jina | 0.60 | 0.75 | 0.54 | 0.75 |

*P=precision, R=recall.*